\documentclass[copyright,creativecommons,fleqn]{eptcs}

\usepackage{amsmath,amssymb}
\usepackage{times,helvet,courier}
\usepackage[vlined,noline]{algorithm2e}\DontPrintSemicolon\SetAlCapSkip{-5pt}\SetAlgoSkip{0pt} 
\usepackage{listings}
\usepackage{multicol}
\usepackage{tikz}
\usepackage{booktabs}
\usepackage[labelfont=bf]{caption}

\usetikzlibrary{arrows,matrix,calc,shapes,positioning,fit}

\newcommand{\aspcud}{\textit{aspcud}}

\newcommand{\clasp}{\textit{clasp}}

\newcommand{\gringo}{\textit{gringo}}

\newcommand{\cudf}{CUDF}

\newcommand{\code}[1]{\text{\lstinline{#1}}}

\newcommand{\targets}[1]{\ensuremath{\mathit{targets}(#1)}}
\newcommand{\Targets}[1]{\ensuremath{\mathit{Targets}(#1)}}
\newcommand{\Provide}[1]{\ensuremath{\mathit{Provide}(#1)}}
\newcommand{\Count}{r}

\SetKwRepeat{Loop}{repeat}{\nl{}until}{}

\newcommand{\squeezy}{\\}
\newcommand{\squeezyend}{\\[-6mm]\nonumber}

\title{\textit{aspcud}: A Linux Package Configuration Tool Based on\\ Answer Set Programming}

\author{Martin Gebser \quad Roland Kaminski \quad Torsten Schaub\thanks{Affiliated with the
                              School of Computing Science at
                              Simon Fraser University,
                              Burnaby, 
                              Canada,
                              and the
                              Institute for Integrated and Intelligent Systems
                              at
                              Griffith University,
                              Brisbane, 
                              Australia.}
  \institute{Universit\"at Potsdam, Institut f\"ur Informatik}
  }
\providecommand{\event}{LoCoCo 2011}
\providecommand{\titlerunning}{\textit{aspcud}: A Linux Package Configuration Tool Based on Answer Set Programming}
\providecommand{\authorrunning}{Martin Gebser, Roland Kaminski, Torsten Schaub}

\begin{document}

\maketitle

\begin{abstract}
We present the Linux package configuration tool \aspcud\
based on Answer Set Programming. 
In particular,
we detail \aspcud's preprocessor turning a \cudf\ specification
into a set of logical facts.
\end{abstract}


\section{Introduction}\label{sec:introduction}

Answer Set Programming (ASP; \cite{baral02a}) owes its increasing popularity as
a tool for Knowledge Representation and Reasoning (KRR;~\cite{KRHandbook}) to
its attractive combination of a rich yet simple modeling language with
high-performance solving capacities.
The basic idea of ASP is to represent a given computational problem by a logic
program whose answer sets correspond to solutions, and then use an ASP solver
for finding answer sets of the program.
This approach is closely related to the one pursued in propositional
Satisfiability Testing (SAT;~\cite{SATHandbook}), where a given problem is encoded as a
propositional theory such that models represent solutions to the problem.
Even though, syntactically, ASP programs resemble Prolog programs, they are
treated by rather different computational mechanisms, based on advanced
Boolean Constraint Satisfaction technology.
Albeit   SAT and ASP both focus on the generation of propositional models,
they differ regarding the semantics of negation, which is
classical in SAT and by default in ASP.
The built-in completion of ``negative knowledge''
admits compact problem  specifications in ASP,
using rules to describe the formation of solution candidates and
integrity constraints to deny unintended ones.

\begin{figure}[t]
\centering\begin{tikzpicture}[
	node distance=0.4cm,
	mycontainer/.style={draw=gray, inner sep=1ex},
	typetag/.style={draw=gray, inner sep=1ex},
	title/.style={draw=none, inner sep=0pt},
	putbelow/.style={below=of #1.south west,anchor=north west,yshift=0.4cm-1ex}]

	\node[title]                   (aspcud) {\aspcud};
	\node[typetag,putbelow=aspcud] (pre)    {Preprocessor};
	\node[typetag,putbelow=pre]    (enc)    {Encoding(s)};
	\node[typetag,right=of pre]    (grd)    {Grounder};
	\node[typetag,right=of grd]    (slv)    {Solver};
	\node[typetag,right=of slv]    (sol)    {Solution};
	\node[typetag,left=of pre]     (in)     {CUDF};

	\path[->] (in)       edge              (pre) ;
	\path[->] (pre)      edge              (grd) ;
	\path[->] (enc.east) edge [bend right] (grd.south) ;
	\path[->] (grd)      edge              (slv) ;
	\path[->] (slv)      edge              (sol) ;

	\node[mycontainer, fit={(aspcud) (pre) (enc) (grd) (slv)}] {};
\end{tikzpicture}
%
\caption{Workflow of \aspcud.\label{fig:architecture}}
\end{figure}
Pioneering work on Linux package configuration was done by Tommi Syrj{\"a}nen
in~\cite{
syrjanen00a}, using ASP for representing and solving
configuration problems for the Debian GNU/Linux system.
%
%
Following this tradition,
we developed the ASP-based Linux package configuration tool \aspcud,
leveraging modern ASP technology for solving 
package configuration
problems posed in the context of the {\small\textsf{mancoosi}} project~\cite{mancoosi}.
As 
shown in Figure~\ref{fig:architecture},
\aspcud\ comprises four components, all of which are freely
available at~\cite{aspcud} (and via~\cite{potassco}).
A given specification (in \cudf;~\cite{trezac09a}) is first preprocessed and mapped to a set of
(logical) facts; this step is explained in Section~\ref{sec:approach}.
As detailed in Section~\ref{sec:system},
the   facts are then combined with one or more (first-order) ASP encodings of
the package configuration problem 
and jointly passed to the ASP grounder \gringo~\cite{gekakosc11a}.
(Our ASP encodings, which are also presented in a companion paper~\cite{gekakasc11c}
 detailing multi-criteria optimization capacities of the ASP solver \clasp~\cite{gekanesc07a}
 and evaluating them on package configuration problems,
 are      provided here for completeness.)
The instantiation of first-order variables upon
grounding 
results in a propositional logic program whose answer sets,
representing problem solutions,
are in turn computed by \clasp.
The impact of
preprocessing on residual problem size as well as solving efficiency
is empirically assessed in Section~\ref{sec:experiments}.
(We do not vary solving strategies here;
 an experimental comparison between different solving strategies
 can be found in~\cite{gekakasc11b,gekakasc11c}.)
Finally,
in Section~\ref{sec:discussion}, we discuss and compare our methodology
with related 
package configuration approaches.
%



\section{Preprocessing}\label{sec:approach}

Our package configuration tool \aspcud\ 
accepts input in
Common Upgradability Description Format (\cudf),
developed in the {\small\textsf{mancoosi}} project
to specify interdependencies of packages
belonging to large software distributions.
The task of a package manager is to find admissible installations satisfying
particular user requests, typically also taking into account soft criteria,
such as minimal change of an existing installation.
While \cudf\ admits arithmetic expressions, package formulae, and virtual packages (see below),
\aspcud's preprocessor generates a flat representation of package interdependencies,
so that they can be conveniently handled by the ASP components of \aspcud\ taking over afterwards.
Below, we       give a quick overview of \cudf\ and optimization criteria, and then
describe the generation of ASP facts.

\subsection{Common Upgradability Description Format (\cudf)}

The general schema of a ``\cudf\ document'' (with an optional \code{preamble}; cf.~\cite{trezac09a}) is as follows:
\begin{lstlisting}[numbers=none,escapechar=?]
 preamble
 package: name?$_1$?    package: name?$_2$?  ?\dots?  package: name?$_n$?
 version: vers?$_1$?    version: vers?$_2$?  ?\dots?  version: vers?$_n$?     request:
 description?$_1$?      description?$_2$?    ?\dots?  description?$_n$?       description
\end{lstlisting}
The pairs $(\code{name}_l,\code{vers}_l)$ for $1\leq l\leq n$
identify installable packages along with positive integer versions; they must be
mutually distinct, that is, $\code{name}_l\neq\code{name}_m$ or $\code{vers}_l\neq\code{vers}_m$
must hold for all $1\leq l < m \leq n$.
Then, the \emph{universe} described by a \cudf\ document is the set
$\mathcal{U}=\{(\code{name}_1,\code{vers}_1),\linebreak[1](\code{name}_2,\code{vers}_2),\linebreak[1]\dots,\linebreak[1](\code{name}_n,\code{vers}_n)\}$
of pairs identifying installable versioned packages.

Each pair $(\code{name}_l,\code{vers}_l)$ can be accompanied with (optional) properties
provided in \code{description}$_l$.
In the most general form, a statement in \code{description}$_l$ looks as follows:
\begin{lstlisting}[numbers=none,escapechar=?]
 property: ?$p_{1_1}$?|?$p_{2_1}$?|?$\dots$?|?$p_{k_1}$?, ?$p_{1_2}$?|?$p_{2_2}$?|?$\dots$?|?$p_{k_2}$?, ?$\dots$?, ?$p_{1_m}$?|?$p_{2_m}$?|?$\dots$?|?$p_{k_m}$?
\end{lstlisting}
In such a statement,
$\code{property}\in\{\code{conflicts},\code{depends},\code{recommends},\code{provides}\}$
determines a kind of package interdependency,
`\code{|}' and `\code{,}' stand for disjunction and conjunction, respectively,
and $p_{j_i}$ for $1\leq i\leq m,1\leq j_i\leq k_i$ is an expression of the form
`\code{name} \code{[op} \code{n]}', in which 
$\code{op}\in\{\code{=},\code{!=},\code{<},\code{<=},\code{>},\code{>=}\}$
denotes an (optional) arithmetic operation along with a positive integer $\code{n}$.
Moreover, if `\code{installed:} \code{true}' is provided in
\code{description}$_l$ for $1\leq l\leq n$,
it means that package \code{name}$_l$ in version \code{vers}$_l$ belongs to an
\emph{existing installation}, and we denote the set of all such pairs $(\code{name}_l,\code{vers}_l)$
by $\mathcal{O}$.

For a $\code{property}\in\{\code{install},\code{remove},\code{upgrade}\}$
in the \code{description} below the keyword `\code{request:}',
for uniformity, we assume the same syntax as with package \code{property} statements 
considered before.\footnote{%
The specification of \cudf~\cite{trezac09a} is more restrictive
by not allowing for disjunction in package formulae associated with 
$\code{property}\in\{\code{conflicts},\code{provides},\code{install},\code{remove},\code{upgrade}\}$.
Moreover, note that \cudf\ additionally admits \code{keep} as  \code{property} in
\code{description}$_l$ for $1\leq l\leq n$,
which we omitted here because it is straightforward to map \code{keep}
to \code{install}.}
The requested properties describe goals that must be satisfied by
a \emph{follow-up installation}~$\mathcal{P}$,
   where certain versioned packages might have to be installed, removed, or
upgraded, respectively.

In order to abstract from arithmetic expressions admitted in \cudf,
for `\code{name} \code{[op} \code{n]}', we define: 
\begin{equation*}
\targets{\code{name}\ \code{[op}\ \code{n]}} = 
  \left\{
  \begin{array}{ll}
    \{(\code{name},n) \mid n\in\mathbb{N}^+ \text{ such that } (n \text{ \code{op} \code{n}}) \text{ holds} \} 
      & \text{if \code{op} \code{n} is specified}
  \\
    \{(\code{name},n) \mid n\in\mathbb{N}^+\} 
      & \text{if \code{op} \code{n} is omitted}
  \end{array}
  \right.
\end{equation*}
We extend the notion of targets to package formulae associated with some
$\code{property}\in\{\code{conflicts},\linebreak[1]\code{depends},\linebreak[1]\code{recommends},\linebreak[1]\code{provides},\linebreak[1]
                     \code{install},\linebreak[1]\code{remove},\linebreak[1]\code{upgrade}\}$
by defining the following multiset:\footnote{%
Multisets are needed to reflect optimization criteria
dealing with (un)satisfied recommendations, below collected in $\mathbf{R}_\mathcal{U}^\mathcal{P}$.}
\begin{equation*}
\Targets{\code{property}} = 
  \left[
  \targets{p_{1_i}}\cup\targets{p_{2_i}}\cup\dots\cup\targets{p_{k_i}}
  \mid
    1\leq i\leq m
  \right]
\end{equation*}
Moreover, let $\Targets{\code{name}_l,\code{vers}_l,\code{property}}$ be
$\Targets{\code{property}}$ for 
$(\code{name}_l,\code{vers}_l)\in\mathcal{U}$ and
$\code{property}\in\{\code{conflicts},\linebreak[1]\code{depends},\linebreak[1]\code{recommends},\linebreak[1]\code{provides}\}$,
where either a unique package formula is provided for \code{property} in \code{description}$_l$,
or $\Targets{\code{property}}=\emptyset$ if \code{property} is not specified in \code{description}$_l$.
Likewise, we let $\Targets{\code{property}}=\emptyset$ for
$\code{property}\in\{\code{install},\linebreak[1]\code{remove},\linebreak[1]\code{upgrade}\}$
if no corresponding statement is provided in the \code{description} below `\code{request:}',
while the package formula defining \code{property} must be unique otherwise.

\begin{figure}[t]
\begin{lstlisting}[multicols=3,numbers=none]
 package:    inst
 version:    3
 conflicts:  conf < 3
 package:    inst
 version:    2
 depends:    dep < 2
 package:    inst
 version:    1
 depends:    dep

 package:    conf
 version:    2
 package:    conf
 version:    1
 installed:  true
 package:    feat
 version:    1
 provides:   conf = 3


 package:    dep
 version:    3
 conflicts:  dep
 recommends: recomm
 package:    dep
 version:    2
 conflicts:  dep < 2
 package:    dep
 version:    1
 installed:  true
 package:    recomm
 version:    1
 conflicts:  option

 package:    option
 version:    1
 depends:    avail

 package:    avail
 version:    1
 installed:  true

 request:    
 install:    inst
 upgrade:    conf > 1
\end{lstlisting}
\caption{\small\cudf\ document specifying the (non-empty) interdependencies
         $\Targets{\code{inst},\code{3},\code{conflicts}}=
          [\{(\code{conf},\code{1}),(\code{conf},\code{2})\}]$,
         $\Targets{\code{inst},\code{2},\code{depends}}=
          [\{(\code{dep},\code{1})\}]$,
         $\Targets{\code{inst},\code{1},\code{depends}}=
          [\{(\code{dep},n)\mid n\in\mathbb{N}\}]$,
         $\Targets{\code{feat},\code{1},\code{provides}}=
          [\{(\code{conf},\code{3})\}]$,
         $\Targets{\code{dep},\code{3},\code{conflicts}}=
          [\{(\code{dep},n)\mid n\in\mathbb{N}\}]$,
         $\Targets{\code{dep},\code{3},\code{recommends}}=
          [\{(\code{recomm},n)\mid n\in\mathbb{N}\}]$,
         $\Targets{\code{dep},\code{2},\code{conflicts}}=
          [\{(\code{dep},\code{1})\}]$,
         $\Targets{\code{recomm},\code{1},\code{conflicts}}=
          [\{(\code{option},n)\mid n\in\mathbb{N}\}]$, and
         $\Targets{\code{option},\code{1},\code{depends}}=
          [\{(\code{avail},n)\mid n\in\mathbb{N}\}]$;
         (non-empty) request targets consist of
         $\Targets{\code{install}}=
          [\{(\code{inst},n)\mid n\in\mathbb{N}\}]$ and
         $\Targets{\code{upgrade}}=
          [\{(\code{conf},n)\mid n\in\mathbb{N},n>1\}]$.\label{cudfinput}}
\end{figure}
As an example, consider the \cudf\ document shown in Figure~\ref{cudfinput}.
The existing installation, marked via `\code{installed:} \code{true}', is
$\mathcal{O}=
 \{
 (\code{conf},\code{1}),\linebreak[1]
 (\code{dep},\code{1}),\linebreak[1]
 (\code{avail},\code{1})
 \}$.
The universe, including all versioned packages,
is 
$\mathcal{U}=\mathcal{O}\cup
 \{
 (\code{inst},\code{3}),\linebreak[1]
 (\code{inst},\code{2}),\linebreak[1]
 (\code{inst},\code{1}),\linebreak[1]
 (\code{conf},\code{2}),\linebreak[1]
 (\code{feat},\code{1}),\linebreak[1]
 (\code{dep},\code{3}),\linebreak[1]
 (\code{dep},\code{2}),\linebreak[1]
 (\code{recomm},\code{1}),\linebreak[1]
 (\code{option},\code{1})
 \}$.
The \cudf\ document further specifies the (non-empty) multisets of targets
of package interdependencies and requests, respectively, provided in the
caption of Figure~\ref{cudfinput};
their particular meanings are described below in the context of ASP fact
generation.

\subsection{Optimization Criteria}

The preprocessor of \aspcud\ takes \emph{optimization criteria}
evaluated in 
competitions by {\small\textsf{mancoosi}}~\cite{mancoosi}
into account.
Given a universe~$\mathcal{U}$, an existing installation~$\mathcal{O}$,
and a follow-up installation~$\mathcal{P}$,
such criteria rely on the minimization or maximization of the following sets:
\begin{eqnarray*}
\mathbf{N}_\mathcal{O}^\mathcal{P} & = & \{ \code{name} \mid (\code{name},\code{vers})\in \mathcal{P},\{(\code{name},n)\mid n\in\mathbb{N}\}\cap \mathcal{O}=\emptyset \} 
\\
\mathbf{D}_\mathcal{O}^\mathcal{P} & = & \{ \code{name} \mid (\code{name},\code{vers})\in \mathcal{O}, \{(\code{name},n)\mid n\in\mathbb{N}\}\cap \mathcal{P}=\emptyset \} 
\\
\mathbf{C}_\mathcal{O}^\mathcal{P} & = & \{ \code{name} \mid (\code{name},\code{vers})\in(\mathcal{P}\setminus \mathcal{O})\cup(\mathcal{O}\setminus \mathcal{P}) \} 
\\
\mathbf{U}_\mathcal{U}^\mathcal{P} & = & \{ \code{name} \mid (\code{name},\code{vers})\in \mathcal{P}, 
(\code{name},\max\{n\mid(\code{name},n)\in\mathcal{U}\})\notin\mathcal{P} \} 
\\
\mathbf{R}_\mathcal{U}^\mathcal{P} & = & \{ (\code{name},\code{vers},i) \mid {} 
\begin{array}[t]{@{}l@{}}
(\code{name},\code{vers})\in \mathcal{P},R_i\cap\Provide{\mathcal{P}}=\emptyset,
{} \\
\Targets{\code{name},\code{vers},\code{recommends}}=[R_1,\dots,R_i,\dots,R_m]\}
\end{array}
\end{eqnarray*}
Here, $\mathbf{N}_\mathcal{O}^\mathcal{P}$ is the collection of packages \code{name}
such that some version \code{vers} 
belongs to~$\mathcal{P}$,
while~$\mathcal{O}$ contains no pair $(\code{name},n)$; that is,
  package \code{name} is new in the follow-up installation~$\mathcal{P}$.
Similarly, $\mathbf{D}_\mathcal{O}^\mathcal{P}$ and $\mathbf{C}_\mathcal{O}^\mathcal{P}$ collect
packages \code{name} that are deleted or changed, respectively,
where change means that some version \code{vers} of \code{name} is new or deleted
in the transition from~$\mathcal{O}$ to~$\mathcal{P}$.
The sets~$\mathbf{U}_\mathcal{U}^\mathcal{P}$ and~$\mathbf{R}_\mathcal{U}^\mathcal{P}$ investigate
the follow-up installation~$\mathcal{P}$ relative to the universe~$\mathcal{U}$.
A package \code{name} belongs to~$\mathbf{U}_\mathcal{U}^\mathcal{P}$ if,
for each pair $(\code{name},\code{vers})$ in~$\mathcal{P}$, 
there is some $(\code{name},n)$ in~$\mathcal{U}$
such that $\code{vers}<n$; that is,
the latest version of \code{name} is missing in~$\mathcal{P}$.
Finally, a triple $(\code{name},\code{vers},i)$ in~$\mathbf{R}_\mathcal{U}^\mathcal{P}$ points to a
disjunction `$p_{1_i}$\code{|}$p_{2_i}$\code{|}$\dots$\code{|}$p_{k_i}$'
in the \code{recommends} statement associated with $(\code{name},\code{vers})$ 
such that $\mathcal{P}$ neither contains nor provides any element of 
$\targets{p_{1_i}}\cup\targets{p_{2_i}}\cup\dots\cup\targets{p_{k_i}}$.
In fact, by $\Provide{\mathcal{P}}=\bigcup_{(\code{name},\code{vers})\in\mathcal{P}}\Provide{\code{name},\code{vers}}$
and $\Provide{\code{name},\code{vers}}=\{(\code{name},\code{vers})\}\cup(\bigcup_{P\in\Targets{\code{name},\code{vers},\code{provides}}}P)$,
we refer to the union of~$\mathcal{P}$ and the targets of
its packages' \code{provides} statements.
This allows us to abstract from ``virtual packages'' that may not be installable themselves,
but can be provided by other packages.
Note that installable and virtual packages are not necessarily disjoint;
e.g., the \cudf\ document in Figure~\ref{cudfinput} specifies
version~\code{1} and~\code{2} of \code{conf} as installable,
while version~\code{3} is provided by $(\code{feat},\code{1})$.
In the following, 
we indicate the objective of \emph{maximizing} or \emph{minimizing}
the \emph{cardinality} of any of the sets~$\mathbf{O}_{\mathcal{O}/\mathcal{U}}^\mathcal{P}$ defined above
by writing $\mathbf{+O}_{\mathcal{O}/\mathcal{U}}^\mathcal{P}$ or $\mathbf{-O}_{\mathcal{O}/\mathcal{U}}^\mathcal{P}$, respectively.

\subsection{Generation of ASP Facts}

We are now ready to specify the algorithm applied by
\aspcud's preprocessor to compute the transitive closure~$\mathcal{C}$ of versioned packages that
may belong to a follow-up installation~$\mathcal{P}$.
The general idea is to include versioned packages by need, that is,
if they are among the targets of some 
\code{install} or \code{upgrade} request, a \code{depends} statement, or
may otherwise serve some user-specified objective.
(E.g., $\mathbf{+N}_\mathcal{O}^\mathcal{P}$
 describes the objective of installing as many new packages as possible,
 so that all pairs $(\code{name},\code{vers})$ in~$\mathcal{U}$ 
 such that \code{name} does not occur in $\mathcal{O}$ would be added to~$\mathcal{C}$.)
Given a universe~$\mathcal{U}$, an existing installation~$\mathcal{O}$, and
a set $\mathbf{O}\subseteq\{
\mathbf{+N}_\mathcal{O}^\mathcal{P},\mathbf{-N}_\mathcal{O}^\mathcal{P},\linebreak[1]
\mathbf{+D}_\mathcal{O}^\mathcal{P},\mathbf{-D}_\mathcal{O}^\mathcal{P},\linebreak[1]
\mathbf{+C}_\mathcal{O}^\mathcal{P},\mathbf{-C}_\mathcal{O}^\mathcal{P},\linebreak[1]
\mathbf{+U}_\mathcal{U}^\mathcal{P},\mathbf{-U}_\mathcal{U}^\mathcal{P},\linebreak[1]
\mathbf{+R}_\mathcal{U}^\mathcal{P},\mathbf{-R}_\mathcal{U}^\mathcal{P}\}$ of objectives,
the transitive closure~$\mathcal{C}$ is computed via 
Algorithm~\ref{algo:cudf2lp}.

In Line~1 of Algorithm~\ref{algo:cudf2lp}, ``negative'' requests given by \code{remove} and also \code{upgrade}
are evaluated; packages that must not be installed are collected in $\mathit{Out}$
to exclude their addition to~$\mathcal{C}$ in the sequel.
While exclusions due to \code{remove} statements are straightforward
(any package fulfilling some \code{remove} target must not be installed),
the issue becomes more involved with \code{upgrade}.
On the one hand, any element of $\Targets{\code{upgrade}}$ resembles an \code{install} request
because it must be served by some package (directly or via a provided virtual package)
in a follow-up installation~$\mathcal{P}$.
On the other hand, there are three additional requirements,
which can make the installation of particular packages prohibitive.
First, the version number of packages subject to \code{upgrade} must 
in a follow-up installation~$\mathcal{P}$ not be smaller than in the existing installation~$\mathcal{O}$
(if some version is provided by~$\mathcal{O}$).
Second, exactly one version must be available in~$\mathcal{P}$,
so that packages providing several versions at once cannot belong to~$\mathcal{P}$.
Third, the \code{install} request implied by an \code{upgrade} target along with
the unique version requirement prohibit the installation of packages providing only
non-matching versions.
These three conditions are taken into account to reflect \code{upgrade} requests
in $\mathit{Out}$.\footnote{%
The \cudf\ specification~\cite{trezac09a} disallows disjunction in \code{upgrade} requests,
and we here generalize \code{upgrade} targets to disjunction in an ``arbitrary'' way.
However, in the case without disjunction,
the packages included in $\mathit{Out}$ due to an \code{upgrade} target
cannot belong to a follow-up installation~$\mathcal{P}$ according to the semantics given in~\cite{trezac09a}.}
(%
For the \cudf\ document in Figure~\ref{cudfinput},
$(\code{conf},\code{2})$ and $(\code{feat},\code{1})$ can fulfill the target
of the \code{upgrade} request `\code{conf} \code{>} \code{1}',
while $(\code{conf},\code{1})$ is excluded in view of its non-matching version.)
Given the set $\mathit{Out}$ of packages that must not belong to a follow-up installation~$\mathcal{P}$,
the test in Line~2 of Algorithm~\ref{algo:cudf2lp} identifies cases in which
\code{install} or \code{upgrade} targets remain unsatisfiable,
regardless of further preprocessing,
so that $\emptyset$ can be immediately returned.
\SetAlgoVlined
\begin{algorithm}[t]
\small
\nl
$\mathit{Out}
 \begin{array}[t]{@{}r@{}l@{}l}
 {} \leftarrow {} &
 \{(\code{name},\code{vers})\in\mathcal{U}
   \mid {} &
   D\in\Targets{\code{remove}},
   D\cap \Provide{\code{name},\code{vers}}\neq\emptyset\}
 \\
 {} \cup {} &
 \{(\code{name},\code{vers})\in\mathcal{U}
   \mid {} & 
   U\in\Targets{\code{upgrade}},(\code{name}',m)\in U,
 {} \\ &
  \multicolumn{2}{l}{
   (\code{name}',n)\in \Provide{\code{name},\code{vers}},
   (\code{name}',n')\in \Provide{\mathcal{O}}, n<n'
  \}}
 \\
 {} \cup {} &
 \{(\code{name},\code{vers})\in\mathcal{U}
   \mid {} &
   U\in\Targets{\code{upgrade}},
 {} \\ &
  \multicolumn{2}{l}{
   1<|\{(\code{name}',n)\in \Provide{\code{name},\code{vers}}\mid(\code{name}',m)\in U\}|
 \}}
 \\
 {} \cup {} &
 \{(\code{name},\code{vers})\in\mathcal{U}
   \mid {} & 
   U\in\Targets{\code{upgrade}},
   U\cap \Provide{\code{name},\code{vers}}=\emptyset,
 {} \\ &
  \multicolumn{2}{l}{
   \{\code{name}' \mid (\code{name}',m)\in U\}\cap
   \{\code{name}' \mid (\code{name}',n)\in \Provide{\code{name},\code{vers}}\}\neq\emptyset
  \}}
 \end{array}$\;
\nl
\lIf{$\{I\in\Targets{\code{install}}\cup\Targets{\code{upgrade}}
        \mid I\cap\Provide{\mathcal{U}\setminus\mathit{Out}}=\emptyset\}\neq\emptyset$}{\Return $\emptyset$}\;
\BlankLine
\nl
$\mathcal{C}\leftarrow
  \{(\code{name},\code{vers})\in\mathcal{U}\setminus\mathit{Out}
    \mid 
    I\in\Targets{\code{install}}\cup\Targets{\code{upgrade}},
    I\cap\Provide{\code{name},\code{vers}}\neq\emptyset\}\!\!\!
$\;
\nl
\lIf{$\mathbf{+N}_\mathcal{O}^\mathcal{P}\in\mathbf{O}$}{%
  $\mathcal{C}\leftarrow\mathcal{C}\cup\{(\code{name},\code{vers})\in\mathcal{U}\setminus\mathit{Out}
   \mid \{n \mid (\code{name},n)\in\mathcal{O}\}=\emptyset\}$}\;
\nl
\lIf{$\mathbf{-D}_\mathcal{O}^\mathcal{P}\in\mathbf{O}$}{%
  $\mathcal{C}\leftarrow\mathcal{C}\cup\{(\code{name},\code{vers})\in\mathcal{U}\setminus\mathit{Out}
   \mid \{n \mid (\code{name},n)\in\mathcal{O}\}\neq\emptyset\}$}\;
\nl
\lIf{$\mathbf{+C}_\mathcal{O}^\mathcal{P}\in\mathbf{O}$}{%
  $\mathcal{C}\leftarrow\mathcal{C}\cup\{(\code{name},\code{vers})\in\mathcal{U}\setminus\mathit{Out}
   \mid (\code{name},\code{vers})\notin\mathcal{O}\}$}\;
\nl
\lIf{$\mathbf{-C}_\mathcal{O}^\mathcal{P}\in\mathbf{O}$}{%
  $\mathcal{C}\leftarrow\mathcal{C}\cup\{(\code{name},\code{vers})\in\mathcal{U}\setminus\mathit{Out}
   \mid (\code{name},\code{vers})\in\mathcal{O}\}$}\;
\nl
\lIf{$\mathbf{+U}_\mathcal{U}^\mathcal{P}\in\mathbf{O}$}{%
  $\mathcal{C}\leftarrow\mathcal{C}\cup\{(\code{name},\code{vers})\in\mathcal{U}\setminus\mathit{Out}
   \mid 
   \code{vers} < \max\{n\mid(\code{name},n)\in\mathcal{U}\}\}$}\;
\nl
\lIf{$\mathbf{+R}_\mathcal{U}^\mathcal{P}\in\mathbf{O}$}{%
  $\mathcal{C}\leftarrow\mathcal{C}\cup\{(\code{name},\code{vers})\in\mathcal{U}\setminus\mathit{Out}
   \mid \Targets{\code{name},\code{vers},\code{recommends}}\neq\emptyset\}$\!\!\!}\;
\BlankLine
\nl
\Loop{$\mathit{Add}=\emptyset$}{%
\nl
$\mathit{Add}\leftarrow
 {} \begin{array}[t]{@{}l@{}}
     \{(\code{name},\code{vers})\in\mathcal{U}\setminus(\mathit{Out}\cup\mathcal{C})
  \mid (\code{name}',\code{vers}')\in\mathcal{C},
 {} \\ \multicolumn{1}{l}{
       D\in\Targets{\code{name}',\code{vers}',\code{depends}},
       D\cap\Provide{\code{name},\code{vers}}\neq\emptyset\}}
 \end{array}$\;
\nl
\lIf{$\mathbf{-R}_\mathcal{U}^\mathcal{P}\in\mathbf{O}$}{%
$\mathit{Add}\leftarrow
 {} \begin{array}[t]{@{}l@{}}
 \mathit{Add}\cup
     \{(\code{name},\code{vers})\in\mathcal{U}\setminus(\mathit{Out}\cup\mathcal{C})
  \mid (\code{name}',\code{vers}')\in\mathcal{C},
 {} \\ \multicolumn{1}{l@{}}{
       R\in\Targets{\code{name}',\code{vers}',\code{recommends}},
       R\cap\Provide{\code{name},\code{vers}}\neq\emptyset\}}
 \end{array}$}\;
\nl
\lIf{$\mathbf{-U}_\mathcal{U}^\mathcal{P}\in\mathbf{O}$}{%
$\mathit{Add}\leftarrow\mathit{Add}\cup
 {} \begin{array}[t]{@{}l@{}} 
     \{(\code{name},\max\{n\mid(\code{name},n)\in\mathcal{U}\})\in\mathcal{U}\setminus(\mathit{Out}\cup\mathcal{C})
  \mid
 {} \\ \multicolumn{1}{l}{
       (\code{name},\code{vers})\in\mathcal{C}\}}
 \end{array}
$}\;
\nl
$\mathcal{C}\leftarrow\mathcal{C}\cup\mathit{Add}$
}
\BlankLine
\nl\Return $\mathcal{C}$
\BlankLine
\BlankLine
\caption{
         Compute transitive closure~$\mathcal{C}$ wrt.\ universe~$\mathcal{U}$,
         existing installation~$\mathcal{O}$, and
         objectives~$\mathbf{O}$.\label{algo:cudf2lp}}
\end{algorithm}

Provided that the test in Line~2 failed,
packages not in $\mathit{Out}$ that may serve some \code{install} or \code{upgrade} target 
are used to initialize the transitive closure~$\mathcal{C}$ in Line~3.
In Line~4--9, $\mathcal{C}$ is further extended in view of the objectives in~$\mathbf{O}$.
As already mentioned, it might be desirable to install any version of a
package \code{name} not occurring in the existing installation~$\mathcal{O}$
if $\mathbf{+N}_\mathcal{O}^\mathcal{P}$ belongs to~$\mathbf{O}$, 
describing the objective of installing as many new packages as possible;
if so, $\mathcal{C}$ is extended accordingly in Line~4.
Note that the objectives of the form $\mathbf{+O}_{\mathcal{O}/\mathcal{U}}^\mathcal{P}$
are useless in practice, as they favor follow-up installations~$\mathcal{P}$ that
are as different from~$\mathcal{O}$, or as suboptimal regarding latest versions or
\code{recommends} targets as possible.
However, such ``anti-optimization'' would in principle be allowed in the \textit{user} track
of competitions by {\small\textsf{mancoosi}}, 
and thus Algorithm~\ref{algo:cudf2lp} includes cases to extend~$\mathcal{C}$ accordingly.
The reasonable cases in Line~5 and~7 apply if package removals or changes, respectively,
are to be minimized,
so that it may help to add all (installed) versions of packages \code{name} occurring
in~$\mathcal{O}$ 
to~$\mathcal{C}$.
For instance, if $\mathbf{-D}_\mathcal{O}^\mathcal{P}$,
aiming at the minimization of package removals,
belongs to~$\mathbf{O}$,
$(\code{conf},\code{2})$, $(\code{dep},\code{3})$, $(\code{dep},\code{2})$, $(\code{dep},\code{1})$,
and $(\code{avail},\code{1})$ are added to~$\mathcal{C}$ in Line~5
for the \cudf\ document in Figure~\ref{cudfinput},
given that $(\code{conf},\code{1})$, $(\code{dep},\code{1})$, and $(\code{avail},\code{1})$
are installed in~$\mathcal{O}$.
Note that the installed pair $(\code{conf},\code{1})$
is not added to~$\mathcal{C}$, as $(\code{conf},\code{1})$ belongs to $\mathit{Out}$.

After its initialization wrt.\ requests (Line~3) and objectives (Line~4--9),
the transitive closure~$\mathcal{C}$ is successively extended in the loop
in Line~10--15 of Algorithm~\ref{algo:cudf2lp}.
To this end, packages $(\code{name},\code{vers})$ matching some dependency of elements
already in~$\mathcal{C}$ are collected in Line~11, provided that the installation 
of $(\code{name},\code{vers})$ is not excluded by  $\mathit{Out}$.
Similarly, packages serving \code{recommends} statements of elements in~$\mathcal{C}$
are collected in Line~12, but only if the minimization of unsatisfied recommendations
is requested via the objective $\mathbf{-R}_\mathcal{U}^\mathcal{P}$.
Finally, if packages ought to be installed in their latest versions,
as it can be specified via $\mathbf{-U}_\mathcal{U}^\mathcal{P}$,
we also collect such latest versions in Line~13.
The three cases justifying the addition of packages to~$\mathcal{C}$
are applied until saturation, and the obtained fixpoint is returned in Line~16.
Any package remaining in $\mathcal{U}\setminus\mathcal{C}$ belongs to $\mathit{Out}$,
meaning that it must not be installed, or is irrelevant regarding dependencies,
requests, and objectives.
Hence, packages outside~$\mathcal{C}$ need not be reflected in ASP facts (described below),
so that both instance and residual problem size can 
be reduced.
For the \cudf\ document in Figure~\ref{cudfinput},
assuming that the objective $\mathbf{-D}_\mathcal{O}^\mathcal{P}$ is provided in~$\mathbf{O}$,
$\mathcal{C}$ is initialized with
\begin{itemize}
\item $(\code{inst},\code{3})$, $(\code{inst},\code{2})$, and $(\code{inst},\code{1})$
      in view of the request `\code{install:} \code{inst}',
\item $(\code{conf},\code{2})$ and $(\code{feat},\code{1})$
      in order to serve `\code{upgrade:} \code{conf} \code{>} \code{1}', and additionally
\item $(\code{dep},\code{3})$, $(\code{dep},\code{2})$, $(\code{dep},\code{1})$,
      and $(\code{avail},\code{1})$ due to the objective~$\mathbf{-D}_\mathcal{O}^\mathcal{P}$.
\end{itemize}
While tracking the dependencies of these packages does not contribute any
further elements to~$\mathcal{C}$, if the objective $\mathbf{-R}_\mathcal{U}^\mathcal{P}$
is given in $\mathbf{O}$, `\code{recommends:} \code{recomm}' associated with $(\code{dep},\code{3})$
justifies the addition of $(\code{recomm},\code{1})$ to~$\mathcal{C}$.
The packages still outside~$\mathcal{C}$ are $(\code{conf},\code{1})$,
which is excluded due to the provided \code{upgrade} request,
and $(\code{option},\code{1})$, as it does not support any element of~$\mathcal{C}$
and could thus be included only if some of the objectives
$\mathbf{+N}_\mathcal{O}^\mathcal{P}$ and $\mathbf{+C}_\mathcal{O}^\mathcal{P}$ would reward
new packages or changes, respectively.

\lstset{keywords=[1]{minimize,maximize,hide,show},keywords=[2]{conflicts,depends,provides,recommends,install,upgrade,installed,keep,unit,newestversion,conflict,request,satisfies,criterion,pconflict,pdepends,psatisfies,in,forbidden,requested,satisfied,violated,opt_define},keywords=[3]{not}}

Given the transitive closure~$\mathcal{C}$ of relevant packages,
the final step of \aspcud's preprocessor is to generate a representation of package
interdependencies, requests, and objectives in terms of ASP facts.
Note that, in competitions by {\small\textsf{mancoosi}}, objectives
are lexicographically ordered by significance;
hence, we below identify~$\mathbf{O}$ with a sequence 
of objectives, written as
$
(\#_1[\mathbf{O}_{\mathcal{O}/\mathcal{U}}^\mathcal{P}]_1,\dots,
              \#_n[\mathbf{O}_{\mathcal{O}/\mathcal{U}}^\mathcal{P}]_n)
$
in increasing order of significance,
where $\#_i\in\{\mathbf{+},\mathbf{-}\}$ and
$[\mathbf{O}_{\mathcal{O}/\mathcal{U}}^\mathcal{P}]_i\in
 \{\mathbf{N}_\mathcal{O}^\mathcal{P},\linebreak[1]
   \mathbf{D}_\mathcal{O}^\mathcal{P},\linebreak[1]
   \mathbf{C}_\mathcal{O}^\mathcal{P},\linebreak[1]
   \mathbf{U}_\mathcal{U}^\mathcal{P},\linebreak[1]
   \mathbf{R}_\mathcal{U}^\mathcal{P}\}$
for $1\leq i\leq n$.
We further associate some ASP constant 
$c_{\mathbf{O}_{\mathcal{O}/\mathcal{U}}^\mathcal{P}}$%
with each $\mathbf{O}_{\mathcal{O}/\mathcal{U}}^\mathcal{P}$
(\code{newpackage} for $\mathbf{O}_\mathcal{O}^\mathcal{P}=\mathbf{N}_\mathcal{O}^\mathcal{P}$,
 \code{remove} for $\mathbf{O}_\mathcal{O}^\mathcal{P}=\mathbf{D}_\mathcal{O}^\mathcal{P}$,
 \code{change} for $\mathbf{O}_\mathcal{O}^\mathcal{P}=\mathbf{C}_\mathcal{O}^\mathcal{P}$,
 \code{uptodate} for $\mathbf{O}_\mathcal{U}^\mathcal{P}=\mathbf{U}_\mathcal{U}^\mathcal{P}$, and
 \code{recommend} for $\mathbf{O}_\mathcal{U}^\mathcal{P}=\mathbf{R}_\mathcal{U}^\mathcal{P}$).
Moreover, for any set $P$ of packages,
we write $\mathit{id}_P$ to refer to some ASP constant
associated with the set~$P$,
where $\mathit{id}_P\neq\mathit{id}_{P'}$ if $P\neq P'$. 
Then, the facts obtained for  a \cudf\ document
(specifying a universe~$\mathcal{U}$ and an existing installation~$\mathcal{O}$),
a sequence~$\mathbf{O}$ of objectives, and~$\mathcal{C}$ are collected in~$\pi$ as 
shown in Figure~\ref{genfact}.%
\begin{figure}[t]
\small%
\vspace*{-4mm}
\begin{eqnarray}
\label{fact:dep}
\tau & = &
\{
\begin{array}[t]{@{}l@{}}
  \code{depends(name,vers,}\mathit{id}_P\code{).}
  \mid
  (\code{name},\code{vers}) \in\mathcal{C},
  D\in\Targets{\code{name},\code{vers},\code{depends}},
{} \\ 
  P = \{(\code{name}',\code{vers}') \in\mathcal{C}
        \mid
        D\cap \Provide{\code{name}',\code{vers}'}\neq\emptyset
      \}
\}
\end{array}
\squeezy\label{fact:rec}
& \cup &
\{
\begin{array}[t]{@{}l@{}}
  \code{recommends(name,vers,}\mathit{id}_P,\Count\code{).}
  \mid
  (\code{name},\code{vers}) \in\mathcal{C},
  \{\mathbf{+R}_\mathcal{U}^\mathcal{P},\mathbf{-R}_\mathcal{U}^\mathcal{P}\}\cap\mathbf{O}\neq\emptyset,
{} \\ 
  \Targets{\code{name},\code{vers},\code{recommends}}=[R_1,\dots,R_i,\dots,R_m], 
{} \\ 
  P = \{(\code{name}',\code{vers}') \in\mathcal{C}
        \mid
        R_i\cap \Provide{\code{name}',\code{vers}'}\neq\emptyset
      \}, 
{} \\ 
  \Count = 
     |\{1\leq j\leq m \mid
        \{
          (\code{name}',\code{vers}') \in\mathcal{C}
          \mid
          R_j\cap \Provide{\code{name}',\code{vers}'}\neq\emptyset
        \}=P
      \}|
\}
\end{array}
\squeezy\label{fact:conf:1}
& \cup &
\{
\begin{array}[t]{@{}l@{}}
  \code{conflict(name,vers,}\mathit{id}_P\code{).}
  \mid
  (\code{name},\code{vers}) \in\mathcal{C},
  C=\bigcup_{T\in\Targets{\code{name},\code{vers},\code{conflicts}}}T,
{} \\ 
  \emptyset\subset
  P = \{(\code{name}',\code{vers}') \in\mathcal{C}\setminus\{(\code{name},\code{vers})\}
        \mid
        C\cap \Provide{\code{name}',\code{vers}'}\neq\emptyset
      \} 
\}
\end{array}
\squeezy\label{fact:conf:2}
& \cup &
\{
\begin{array}[t]{@{}l@{}}
  \code{conflict(name,vers,}\mathit{id}_P\code{).}
  \mid
  (\code{name},\code{vers}) \in\mathcal{C},
{} \\ 
  U\in\Targets{\code{upgrade}},U\cap\Provide{\code{name},\code{vers}}\neq\emptyset,
{} \\
  \emptyset\subset
  P = \{(\code{name}',\code{vers}') \in\mathcal{C}
        \mid
{} \begin{array}[t]{@{}l@{}}
        U\cap \Provide{\code{name}',\code{vers}'}\neq\emptyset,
{} \\
        U\cap \Provide{\code{name}',\code{vers}'} \neq U\cap\Provide{\code{name},\code{vers}}
      \} 
\}
   \end{array}
\end{array}
\squeezy\label{fact:req}
& \cup &
\{
\begin{array}[t]{@{}l@{}}
  \code{request(}\mathit{id}_P\code{).} 
  \mid
  I \in \Targets{\code{install}} \cup \Targets{\code{upgrade}},
{} \\
  P = \{(\code{name},\code{vers})\in\mathcal{C}
        \mid
        I\cap \Provide{\code{name},\code{vers}}\neq\emptyset\}
\}
\end{array}
\\[2mm]\nonumber
\pi & = & \tau
\squeezy\label{fact:sat:1}
& \cup &
\{
\begin{array}[t]{@{}l@{}}
  \code{satisfies(name,vers,}\mathit{id}_P\code{).} 
  \mid
  (\code{name},\code{vers})\in P,
  (\code{depends(name}'\code{,vers}'\code{,}\mathit{id}_P\code{).})\in\tau 
\}
\end{array}
\squeezy\label{fact:sat:2}
& \cup &
\{
\begin{array}[t]{@{}l@{}l@{}}
  \code{satisfies(name,vers,}\mathit{id}_P\code{).} 
  \mid
{} &
  (\code{name},\code{vers})\in P,
{} \\ &
  (\code{recommends(name}'\code{,vers}'\code{,}\mathit{id}_P,\Count\code{).})\in\tau
\}
\end{array}
\squeezy\label{fact:sat:3}
& \cup &
\{
\begin{array}[t]{@{}l@{}}
  \code{satisfies(name,vers,}\mathit{id}_P\code{).} 
  \mid
  (\code{name},\code{vers})\in P,
  (\code{conflict(name}'\code{,vers}'\code{,}\mathit{id}_P\code{).})\in\tau 
\}
\end{array}
\squeezy\label{fact:sat:4}
& \cup &
\{
\begin{array}[t]{@{}l@{}}
  \code{satisfies(name,vers,}\mathit{id}_P\code{).} 
  \mid
  (\code{name},\code{vers})\in P,
  (\code{request(}\mathit{id}_P\code{).})\in\tau
\}
\end{array}
\squeezy\label{fact:unit}
& \cup &
\{
  \code{unit(name,vers).}
  \mid 
  (\code{name},\code{vers}) \in\mathcal{C}
\}
\squeezy\label{fact:ins}
& \cup &
\{
  \code{installed(name,vers).}
  \mid 
  (\code{name},\code{vers}) \in\mathcal{O}
\}
\squeezy\label{fact:new}
& \cup &
\{
  \code{newestversion(name,}\max\{n\mid(\code{name},n)\in\mathcal{U}\}\code{).}
  \mid 
  (\code{name},\code{vers}) \in\mathcal{C}
\}
\squeezy\label{fact:crit}
& \cup &
\{
  \code{criterion(}c_{[\mathbf{O}_{\mathcal{O}/\mathcal{U}}^\mathcal{P}]_i}\code{,}\#_i i\code{).} 
  \mid
  \mathbf{O}=(\#_1[\mathbf{O}_{\mathcal{O}/\mathcal{U}}^\mathcal{P}]_1,\dots,
              \#_n[\mathbf{O}_{\mathcal{O}/\mathcal{U}}^\mathcal{P}]_n),
             1\leq i\leq n
\}
\squeezyend
\end{eqnarray}
\caption{ASP facts for a \cudf\ document,
a sequence~$\mathbf{O}$ of objectives, and a set~$\mathcal{C}\subseteq\mathcal{U}$
of packages.\label{genfact}}
\end{figure}

In Figure~\ref{genfact},
the subset~$\tau$ of~$\pi$ 
groups packages fulfilling targets of package interdependencies or requests in sets~$P$, and
respective facts introduce constants $\mathit{id}_P$ 
referring to~$P$.
While facts over the predicate \code{depends}
in~(\ref{fact:dep}) simply link the targets of dependencies to
packages that provide them,
\code{recommends} in~(\ref{fact:rec})
introduces a counter~$\Count$ along with
each set~$P$ of packages fulfilling a recommendation~$R_i$
because several elements of the multiset
$\Targets{\code{name},\code{vers},\linebreak[1]\code{recommends}}=[R_1,\dots,R_i,\dots,R_m]$
may share the same providers~$P$.
Also note that~(\ref{fact:rec}) contributes facts to~$\tau$ (and~$\pi$)
only if $\#\mathbf{R}_\mathcal{U}^\mathcal{P}$ for $\#\in\{\mathbf{+},\mathbf{-}\}$
is among the objectives in~$\mathbf{O}$.
The packages~$P$ considered by \code{conflict} in~(\ref{fact:conf:1})
are obtained by joining all $T\in\Targets{\code{name},\linebreak[1]\code{vers},\linebreak[1]\code{conflicts}}$
in~$C$ before collecting their providers in~$P$.
Note that $(\code{name},\code{vers})$ can by definition (cf.~\cite{trezac09a})
not be in conflict with itself, even if it fulfills some
$T\in\Targets{\code{name},\linebreak[1]\code{vers},\linebreak[1]\code{conflicts}}$;
this situation arises with $(\code{dep},\code{3})$ in Figure~\ref{cudfinput},
where `\code{conflicts:} \code{dep}'
specifies a universal conflict with any version of \code{dep}
(and packages including \code{dep} in their \code{provides} statements).
Additional conflicts may be induced by \code{upgrade} requests
in view of their unique version requirement, and thus
packages providing different elements of some $U\in\Targets{\code{upgrade}}$
are marked as conflicting via (\ref{fact:conf:2});
for instance,
the \code{upgrade} request `\code{conf} \code{>} \code{1}' in Figure~\ref{cudfinput}
is reflected by facts 
`\code{conflict(conf,2,}\linebreak[1]$\mathit{id}_{\{(\code{feat},\code{1})\}}$\code{).}' and
`\code{conflict(feat,1,}\linebreak[1]$\mathit{id}_{\{(\code{conf},\code{2})\}}$\code{).}',
obtained because $(\code{feat},\code{1})$ provides $(\code{conf},\code{3})$
(as a virtual package).
Finally, facts over the predicate \code{request} in~(\ref{fact:req})
group packages~$P$ fulfilling \code{install} or \code{upgrade} requests
to express that some element of~$P$  must    be included in a follow-up
installation~$\mathcal{P}$.
Note that all packages referred to in facts of~$\tau$,
via $(\code{name},\code{vers})$ in arguments or belonging to
$P$ associated with some constant $\mathit{id}_P$, 
are elements of the transitive closure~$\mathcal{C}$;
that is, the package interdependencies and requests specified by~$\tau$ 
are limited to~$\mathcal{C}$.

\begin{figure}[t]
\begin{lstlisting}[multicols=2,numbers=none,escapechar=?,basicstyle=\ttfamily\small]
unit(inst,3).
conflict(inst,3,?$\mathit{id}_{\{(\text{\texttt{conf}},\text{\texttt{2}})\}}$?).
unit(inst,2).
depends(inst,2,?$\mathit{id}_{\{(\text{\texttt{dep}},\text{\texttt{1}})\}}$?).
unit(inst,1).
depends(inst,1,?$\mathit{id}_{\{(\text{\texttt{dep}},\text{\texttt{3}}),(\text{\texttt{dep}},\text{\texttt{2}}),(\text{\texttt{dep}},\text{\texttt{1}})\}}$?).
newestversion(inst,3).

unit(conf,2).
conflict(conf,2,?$\mathit{id}_{\{(\text{\texttt{feat}},\text{\texttt{1}})\}}$?).
newestversion(conf,2).
installed(conf,1).

unit(feat,1).
conflict(feat,1,?$\mathit{id}_{\{(\text{\texttt{conf}},\text{\texttt{2}})\}}$?).
newestversion(feat,1).

unit(dep,3).
conflict(dep,3,?$\mathit{id}_{\{(\text{\texttt{dep}},\text{\texttt{2}}),(\text{\texttt{dep}},\text{\texttt{1}})\}}$?).
unit(dep,2).
conflict(dep,2,?$\mathit{id}_{\{(\text{\texttt{dep}},\text{\texttt{1}})\}}$?).
unit(dep,1).
newestversion(dep,3).
installed(dep,1).
unit(avail,1).
newestversion(avail,1).
installed(avail,1).

request(?$\mathit{id}_{\{(\text{\texttt{inst}},\text{\texttt{3}}),(\text{\texttt{inst}},\text{\texttt{2}}),(\text{\texttt{inst}},\text{\texttt{1}})\}}$?).
request(?$\mathit{id}_{\{(\text{\texttt{conf}},\text{\texttt{2}}),(\text{\texttt{feat}},\text{\texttt{1}})\}}$?).

satisfies(conf,2,?$\mathit{id}_{\{(\text{\texttt{conf}},\text{\texttt{2}})\}}$?).
satisfies(dep,1,?$\mathit{id}_{\{(\text{\texttt{dep}},\text{\texttt{1}})\}}$?).
satisfies(dep,3,?$\mathit{id}_{\{(\text{\texttt{dep}},\text{\texttt{3}}),(\text{\texttt{dep}},\text{\texttt{2}}),(\text{\texttt{dep}},\text{\texttt{1}})\}}$?).
satisfies(dep,2,?$\mathit{id}_{\{(\text{\texttt{dep}},\text{\texttt{3}}),(\text{\texttt{dep}},\text{\texttt{2}}),(\text{\texttt{dep}},\text{\texttt{1}})\}}$?).
satisfies(dep,1,?$\mathit{id}_{\{(\text{\texttt{dep}},\text{\texttt{3}}),(\text{\texttt{dep}},\text{\texttt{2}}),(\text{\texttt{dep}},\text{\texttt{1}})\}}$?).
satisfies(feat,1,?$\mathit{id}_{\{(\text{\texttt{feat}},\text{\texttt{1}})\}}$?).
satisfies(dep,2,?$\mathit{id}_{\{(\text{\texttt{dep}},\text{\texttt{2}}),(\text{\texttt{dep}},\text{\texttt{1}})\}}$?).
satisfies(dep,1,?$\mathit{id}_{\{(\text{\texttt{dep}},\text{\texttt{2}}),(\text{\texttt{dep}},\text{\texttt{1}})\}}$?).
satisfies(inst,3,?$\mathit{id}_{\{(\text{\texttt{inst}},\text{\texttt{3}}),(\text{\texttt{inst}},\text{\texttt{2}}),(\text{\texttt{inst}},\text{\texttt{1}})\}}$?).
satisfies(inst,2,?$\mathit{id}_{\{(\text{\texttt{inst}},\text{\texttt{3}}),(\text{\texttt{inst}},\text{\texttt{2}}),(\text{\texttt{inst}},\text{\texttt{1}})\}}$?).
satisfies(inst,1,?$\mathit{id}_{\{(\text{\texttt{inst}},\text{\texttt{3}}),(\text{\texttt{inst}},\text{\texttt{2}}),(\text{\texttt{inst}},\text{\texttt{1}})\}}$?).
satisfies(conf,2,?$\mathit{id}_{\{(\text{\texttt{conf}},\text{\texttt{2}}),(\text{\texttt{feat}},\text{\texttt{1}})\}}$?).
satisfies(feat,1,?$\mathit{id}_{\{(\text{\texttt{conf}},\text{\texttt{2}}),(\text{\texttt{feat}},\text{\texttt{1}})\}}$?).

criterion(change,-1).
criterion(remove,-2).
\end{lstlisting}
\caption{ASP facts $\pi$ obtained for the \cudf\ document in Figure~\ref{cudfinput}
         along with $\mathbf{O}=(\mathbf{-C}_{\mathcal{O}}^\mathcal{P},\mathbf{-D}_{\mathcal{O}}^\mathcal{P})$.\label{fig:facts}}
\end{figure}
The full ASP instance~$\pi$ extracted from a \cudf\ document
is obtained by joining~$\tau$ with further facts.
The first group of them, given in~(\ref{fact:sat:1})--(\ref{fact:sat:4}) in Figure~\ref{genfact},
links packages $(\code{name},\code{vers})\in P$ to 
$\mathit{id}_P$ 
via the predicate \code{satisfies}, 
where $\mathit{id}_P$ 
was 
introduced in~$\tau$.
The second group of facts in~(\ref{fact:unit})--(\ref{fact:new})
describes the transitive closure~$\mathcal{C}$,
the existing installation~$\mathcal{O}$,
and latest versions of packages in~$\mathcal{C}$
via the predicates 
\code{unit}, \code{installed}, and \code{newestversion}.
Moreover,     facts over the predicate \code{criterion} in~(\ref{fact:crit})
represent objectives
$\#_i[\mathbf{O}_{\mathcal{O}/\mathcal{U}}^\mathcal{P}]_i$ occurring in~$\mathbf{O}$
by an associated constant
$c_{[\mathbf{O}_{\mathcal{O}/\mathcal{U}}^\mathcal{P}]_i}$ and the
polarity $\#_i\in\{\mathbf{+},\mathbf{-}\}$ along
with the position $i$ in~$\mathbf{O}$.
E.g., the facts obtained for
the \cudf\ document in Figure~\ref{cudfinput} and
the sequence $\mathbf{O}=(\mathbf{-C}_{\mathcal{O}}^\mathcal{P},\mathbf{-D}_{\mathcal{O}}^\mathcal{P})$
of objectives 
are shown in Figure~\ref{fig:facts}.
Note that, in view of unspecified objectives regarding recommendations,
the respective interdependency of package $(\code{dep},\code{3})$
is not reflected in the facts.
However, when $\mathbf{-R}_{\mathcal{U}}^\mathcal{P}$ would be added to~$\mathbf{O}$,
`$\code{recommends(dep,3,}\linebreak[1]\mathit{id}_{\{(\code{recomm},\code{1})\}}\code{,1).}$'
along with further facts describing 
$(\code{recomm},\code{1})$ (then also included in~$\mathcal{C}$)
would be obtained in~$\pi$.%

\section{
Grounding and Solving}\label{sec:system}

The facts~$\pi$ generated by the preprocessor serve as 
problem-specific input to the ASP components of \aspcud,
viz., the grounder \gringo~\cite{gekakosc11a} and the solver \clasp~\cite{gekanesc07a},
while general knowledge about package configuration problems is provided via encodings.
For one, the encoding \code{configuration.lp} in Figure~\ref{fig:configuration}
specifies admissible follow-up installations~$\mathcal{P}$; for another,
\code{optimization.lp} in Figure~\ref{fig:optimize} encodes optimization
criteria (violations) and corresponding penalties.
The encodings are written in the first-order input language of \gringo,
which instantiates the contained variables wrt.~$\pi$ to produce a propositional
representation suitable for \clasp.
For space reasons, we confine the presentation to the encodings
that appeared to be most successful in our preliminary, systematic experiments
and are thus used by default in \aspcud.
However, major strengths of ASP are its first-order input language and the
availability of grounders to instantiate them;
this enables rapid prototyping of alternative problem formulations,
and we indeed tested several encoding variants before deciding for the ones
provided next.

\subsection{Hard Constraints}

Hard requirements for follow-up installations~$\mathcal{P}$ 
are encoded in \code{configuration.lp}.
Here, the rules in Line~3--10 are used to abstract
from versions if a property applies to all (installable) versions
of a package.
Note that variables are universally quantified,
where~\code{P} stands for the name a package,
\code{X} for a version of~\code{P},
and \code{D} is an identifier, $\mathit{id}_P$,
for a set~$P$ of packages.
In view of this, the auxiliary predicate \code{pconflict} defined
in Line~3 projects out versions~\code{X} from facts
over \code{conflict} in~$\pi$.
The rule in Line~4 then lifts a conflict between some version of~\code{P}
(and packages fulfilling~\code{D}) to the package name~\code{P}, provided that
   all (installable) versions~\code{X} 
conflict with~\code{D};
in fact, the condition `\code{conflict(P,X,D)} \code{:} \code{unit(P,X)}',
evaluated wrt.\ values for~\code{P} and~\code{D} given through \code{pconflict(P,D)},
refers to the conjunction of \code{conflict(P,X,D)} over all instances of~\code{X}
such that \code{unit(P,X)} holds.
From the facts~$\pi$ in Figure~\ref{fig:facts},
\code{conflict(conf,}$\mathit{id}_{\{(\code{feat},\code{1})\}}$\code{)} and
\code{conflict(feat,}$\mathit{id}_{\{(\code{conf},\code{2})\}}$\code{)}
are derived via instances of the rules in Line~3 and~4,
as \code{conflict(conf,2,}$\mathit{id}_{\{(\code{feat},\code{1})\}}$\code{)} and
\code{conflict(feat,1,}$\mathit{id}_{\{(\code{conf},\code{2})\}}$\code{)} are 
provided by facts for the only (installable) versions~\code{2} and~\code{1}
of \code{conf} and \code{feat}, respectively.
The 
         same approach to lift properties to package names~\code{P}
is 
            applied to dependencies and satisfaction relationships (i.e.,
membership in a   set~$P$ referred to by some $\mathit{id}_P$,
given via facts over the predicate \code{satisfies}).%
\begin{figure}[t]
\begin{lstlisting}[aboveskip=0pt,belowskip=-4pt,basicstyle=\ttfamily\small]
% analyze package interdependencies

 pconflict(P,D) :-   conflict(P,X,D).
  conflict(P,D) :-  pconflict(P,  D),  conflict(P,X,D) : unit(P,X).

  pdepends(P,D) :-    depends(P,X,D).
   depends(P,D) :-   pdepends(P,  D),   depends(P,X,D) : unit(P,X).

psatisfies(P,D) :-  satisfies(P,X,D).
 satisfies(P,D) :- psatisfies(P,  D), satisfies(P,X,D) : unit(P,X).

% generate follow-up installation

{ in(P,X) }     :- unit(P,X).
  in(P)         :-   in(P,X).

forbidden(D)    :-   in(P,X),  conflict(P,X,D).
forbidden(D)    :-   in(P),    conflict(P,  D).

requested(D)    :-   in(P,X),   depends(P,X,D).
requested(D)    :-   in(P),     depends(P,  D).

satisfied(D)    :-   in(P,X), satisfies(P,X,D).
satisfied(D)    :-   in(P),   satisfies(P,  D).

 :-   request(D), not satisfied(D).
 :- requested(D), not satisfied(D).
 :- forbidden(D),     satisfied(D).

% project output

#hide.  #show in/2.
\end{lstlisting}
\caption{ASP encoding of follow-up installations~$\mathcal{P}$ wrt.\ facts~$\pi$ (\code{configuration.lp}).\label{fig:configuration}}
\end{figure}%

While the rules described so far derive deterministic properties from facts,
the ``choice'' rule in Line~14 of \code{configuration.lp} allows for guessing a
follow-up installation~$\mathcal{P}$.
It describes that, for any instance of          $(\code{P},\code{X})$
specified by the predicate \code{unit},
one may freely choose whether to include \code{in(P,X)} in an answer set;
and a follow-up installation~$\mathcal{P}$ is given by the instances of
\code{in(P,X)} belonging to an answer set.
Hence, the rule in Line~14 opens up the candidate space for~$\mathcal{P}$,
which is however limited to the transitive closure~$\mathcal{C}$ 
(determined via Algorithm~\ref{algo:cudf2lp}) because facts over
\code{unit} do not include packages outside~$\mathcal{C}$.
The rule in Line~15 again abstracts from the version~\code{X} of
a package~\code{P} in~$\mathcal{P}$ by projecting out~\code{X} from \code{in(P,X)}.
Once guessed,
it remains to check whether a follow-up installation~$\mathcal{P}$ is admissible.
To this end,
the rules in Line~17--24 collect the identifiers  $\mathit{id}_P$  of target sets~$P$
of package interdependencies,
divided by \code{forbidden} and \code{requested} target sets in
view of conflicts and dependencies, respectively, of packages in~$\mathcal{P}$,
and \code{satisfied} target sets are determined in turn.
The actual checks are implemented via the ``constraints'' in Line~26--28,
which deny follow-up installations~$\mathcal{P}$ such that
the target set of a \code{request} (due to some \code{install} or \code{upgrade}
statement in the original \cudf\ document) or a \code{requested} package dependency
is not \code{satisfied}; furthermore, a target set \code{forbidden} in view of some conflict
must not be \code{satisfied}.
For instance,
the requirement expressed by
`\code{request(}$\mathit{id}_{\{(\code{inst},\code{3}),(\code{inst},\code{2}),(\code{inst},\code{1})\}}$\code{).}'
in Figure~\ref{fig:facts} along with the constraint in Line~26 deny
follow-up installations~$\mathcal{P}$ that do not include any of
the packages $(\code{inst},\code{3})$, $(\code{inst},\code{2})$, and $(\code{inst},\code{1})$
because 
\code{satisfied(}$\mathit{id}_{\{(\code{inst},\code{3}),(\code{inst},\code{2}),(\code{inst},\code{1})\}}$\code{)}
can be derived only if \code{in(inst,}$n$\code{)}
holds for some $n\in\{\code{1},\code{2},\code{3}\}$.
If so,
an instance of the rule in Line~23 as well as the rules in Line~15 and~24 apply,
where the latter relies on
\code{satisfies(inst,}$\mathit{id}_{\{(\code{inst},\code{3}),(\code{inst},\code{2}),(\code{inst},\code{1})\}}$\code{)},
which abstracts from versions of \code{inst}.
Note that such abstractions and the rules in Line~18, 21, and~24 exploiting them
are in principle redundant, since analogous rules considering versions in Line~17, 20, and~23
achieve the same effect, once a version~\code{X} of~\code{P} is determined via \code{in(P,X)}.
However, our preliminary empirical comparisons between several encoding variants
suggested \code{configuration.lp} in Figure~\ref{fig:configuration} as the
most ``efficient'' encoding.
Finally, 
an admissible follow-up installation~$\mathcal{P}$
can be read off from instances of \code{in(P,X)} belonging to an answer set,
and so we confine its displayed part accordingly in Line~32.

\subsection{Soft Constraints}

The encoding    \code{optimization.lp} in Figure~\ref{fig:optimize} builds on top of facts~$\pi$ and \code{configuration.lp}
to identify optimization criteria violations and to assign corresponding penalties.
While the rule in Line~1 merely
projects out versions~\code{X} of packages~\code{P} \code{installed} in~$\mathcal{O}$,
the rules in Line~5--12 recognize changes, additions, and removals of packages~\code{P}
in the transition from~$\mathcal{O}$ to~$\mathcal{P}$.
Note that any such \code{violated} maintenance condition is considered only if
 associated objectives are specified via facts over the predicate \code{criterion} in~$\pi$;
for the facts in Figure~\ref{fig:facts}, the rules in Line~5--8 and~11--12 of Figure~\ref{fig:optimize} are applicable, given that
the sequence $\mathbf{O}=(\mathbf{-C}_{\mathcal{O}}^\mathcal{P},\mathbf{-D}_{\mathcal{O}}^\mathcal{P})$ 
of objectives is expressed via `\code{criterion(change,-1).}' and `\code{criterion(remove,-2).}' 
Objectives regarding latest versions of packages in~$\mathcal{P}$ and
recommendations are addressed by the rules in Line~13--14 and 15--16, respectively.
Note that the latter uses a different format, \code{r(P,X,D)}, 
to indicate an unserved recommendation~\code{D}
of a package~\code{P} in version~\code{X}, where~\code{D} is an identifier of the form
$\mathit{id}_P$ for a target set~$P$;
in addition, the multiplicity of recommendation targets served by~$P$ is given in~\code{R}.
(Since     violations of the other optimization criteria, identified in Line~5--14,
are counted once per package name~\code{P},
their corresponding instances of \code{violated(C,P,1)} use~\code{1} as default weight.)
The \code{\#minimize} and \code{\#maximize} statements
in Line~20 and~21 associate penalties (or rewards) with
violations of objectives of the form $\#_i[\mathbf{O}_{\mathcal{O}/\mathcal{U}}^\mathcal{P}]_i$
in a sequence~$\mathbf{O}$,
reflected in~$\pi$ by including
`$\code{criterion(}c_{[\mathbf{O}_{\mathcal{O}/\mathcal{U}}^\mathcal{P}]_i}\code{,}\#_i i\code{).}$'
(where
$c_{[\mathbf{O}_{\mathcal{O}/\mathcal{U}}^\mathcal{P}]_i}\in
 \{
 \code{newpackage},\linebreak[1]
 \code{remove},\linebreak[1]
 \code{change},\linebreak[1]
 \code{uptodate},\linebreak[1]
 \code{recommend}
 \}$
and 
$\#_i\in\{\mathbf{+},\mathbf{-}\}$).
Instances of \code{violated(}$c_{[\mathbf{O}_{\mathcal{O}/\mathcal{U}}^\mathcal{P}]_i}$\code{,P,W)}
in an answer set, derived via the rules in Line~5--16,
are then penalized (or rewarded) with priority~$i$ and weight~\code{W}.
Note that summation-based minimization applies (in Line~20) if $\#_i=\mathbf{-}$  or
maximization (in Line~21) if $\#_i=\mathbf{+}$,
while a later position~$i$ in~$\mathbf{O}$ indicates greater significance than
preceding ones.
For instance, the sequence represented by 
`\code{criterion(change,-1).}' and `\code{criterion(remove,-2).}'
gives preference to the minimization of $\mathbf{D}_\mathcal{O}^\mathcal{P}$ and then
considers the cardinality 
of $\mathbf{C}_\mathcal{O}^\mathcal{P}$ for breaking ties.
As already mentioned,
maximization objectives of the form $\mathbf{+O}_{\mathcal{O}/\mathcal{U}}^\mathcal{P}$
(aiming at many differences between $\mathcal{O}$ and~$\mathcal{P}$,
outdated packages in~$\mathcal{P}$, or
recommendations ignored by~$\mathcal{P}$, respectively)
seem of little practical use.
Since they would still be allowed in the \textit{user} track
of competitions by {\small\textsf{mancoosi}},
the \code{\#maximize} statement in Line~21 of Figure~\ref{fig:optimize}
is included to handle them.%
\begin{figure}[t]
\begin{lstlisting}[aboveskip=0pt,belowskip=-4pt,basicstyle=\ttfamily\small]
installed(P) :- installed(P,X).

% identify optimization criteria violations

violated(change,     P,     1) :- criterion(change,    L),
                              installed(P,X), not in(P,X).
violated(change,     P,     1) :- criterion(change,    L),
                          not installed(P,X),     in(P,X).
violated(newpackage, P,     1) :- criterion(newpackage,L),
                          not installed(P),       in(P).
violated(remove,     P,     1) :- criterion(remove,    L),
                              installed(P),   not in(P).
violated(uptodate,   P,     1) :- criterion(uptodate,  L),
                          newestversion(P,X), not in(P,X), in(P).
violated(recommend,r(P,X,D),R) :- criterion(recommend, L),
                             recommends(P,X,D,R), in(P,X), not satisfied(D).

% post optimization criteria

#minimize[ violated(C,P,W) = W @ -L : criterion(C,L) : L < 0 ].
#maximize[ violated(C,P,W) = W @  L : criterion(C,L) : L > 0 ].
\end{lstlisting}
\caption{ASP encoding of optimization criteria wrt.\ follow-up installations~$\mathcal{P}$ (\code{optimization.lp}).\label{fig:optimize}}
\end{figure}%

The instantiation of \code{configuration.lp} and \code{optimization.lp} wrt.\ 
facts~$\pi$, produced by \gringo, is passed on to the ASP solver \clasp,
which searches for (optimal) answer sets of propositional logic programs.
In the context of Linux package configuration,
the major challenge lies in the optimization of objectives,
given that available distributions are large
and plenty installations are admissible
(even when the transitive closure~$\mathcal{C}$ is used to
 limit the scope of a follow-up installation~$\mathcal{P}$).
In view of this,
we recently extended \clasp\ by dedicated search strategies and heuristics 
for effective multi-criteria optimization~\cite{gekakasc11b};
by default,
\aspcud\ configures them by supplying the command line options
\code{--opt-hierarch=1} and \code{--opt-heuristic=1} to \clasp.
(Default \clasp\ options can be overridden via \aspcud\ switch `\code{-c}'.)
In a nutshell, these options instruct \clasp\ to optimize multiple objectives
successively      in the order of significance by progressively
improving objective values of answer sets until the problem of
finding a better answer set turns out to be unsatisfiable,
in which case optimization proceeds with the next (less significant) criterion.
Further search parameters of \clasp\ 
are, by default, set by supplying the command line options 
\code{--sat-prepro},
\code{--heuristic=vsids},
\code{--solution-recording},
\code{--restarts=128}, and
\code{--local-restarts}.
We determined the \clasp\ setting utilized by \aspcud\ via systematic experiments
(see~\cite{gekakasc11b,gekakasc11c} for an empirical comparison between \clasp\ settings),
and the successful participations of \aspcud\ in recent trial-runs
of the competition by {\small\textsf{mancoosi}}~\cite{mancoosi} 
were largely owed to the search capacities of \aspcud's solving component.

\lstset{keywords=[2]{remove,conflicts,depends,provides,recommends,install,upgrade,installed,keep,unit,newestversion,conflict,request,satisfies,criterion,pconflict,pdepends,psatisfies,in,forbidden,requested,satisfied,violated,opt_define}}

\section{Experiments}\label{sec:experiments}

The workflow of \aspcud\ includes the steps of preprocessing, grounding, and solving
(as well as converting an answer set representing a follow-up installation back to \cudf).
Since \clasp\ settings were already evaluated in~\cite{gekakasc11b,gekakasc11c},
the experiments presented here concentrate on the impact of
preprocessing on residual problem size and its effect on solving efficiency.
%
To be more precise,
we compare problem size and search statistics wrt.\
ASP facts limited to the transitive closure~$\mathcal{C}$ determined via 
Algorithm~\ref{algo:cudf2lp} against
facts describing the whole universe~$\mathcal{U}$ of packages
(except for those that must not be installed in view of
 \code{remove} and \code{upgrade} requests).

Our experiments consider four benchmark classes, in the following referred to by
\textit{easy}, \textit{difficult}, \textit{impossible}, and \textit{debian-dudf},
from the 2010 MISC competition by {\small\textsf{mancoosi}}~\cite{mancoosi}. %
Furthermore, we apply the sequences
$(\mathbf{-C}_{\mathcal{O}}^\mathcal{P},\mathbf{-D}_{\mathcal{O}}^\mathcal{P})$ and
$(\mathbf{-N}_{\mathcal{O}}^\mathcal{P},\mathbf{-R}_{\mathcal{U}}^\mathcal{P},\mathbf{-U}_{\mathcal{U}}^\mathcal{P},\mathbf{-D}_{\mathcal{O}}^\mathcal{P})$
of objectives (in increasing order of significance)
used in the tracks called \textit{paranoid} and \textit{trendy}.
(Arbitrary sequences of objectives can be provided as arguments to \aspcud,
 as required in the \textit{user} track.)
Note that, although the instances are the same in \textit{paranoid} and \textit{trendy} mode,
optimization wrt.\ the latter is usually more difficult in view
of 
more criteria.
%
%
%
%
We ran the experiments under MISC 
conditions,
imposing a time limit of 300 seconds,
on an Intel 
Xeon E5520 machine,
equipped with 2.27GHz processors and 48GB main memory, under Linux.

Table~\ref{tab:experiments} summarizes experimental results,
separately for \textit{paranoid} and \textit{trendy} objectives,
where 
the first two columns provide
the considered benchmark class along with the number~$n$ of its instances.
The entries in the other columns contrast statistics obtained
with transitive closure computation (before~`/')
against the ones obtained without it (after~`/').
Average problem sizes in terms of
number of variables and constraints, as reported by \clasp, 
are provided in the third and fourth column.
The fifth column gives average solving times, with timeouts (in parentheses)
taken as 300 seconds.
The numbers of choices, conflicts,
and answer sets (including intermediate ones) 
reported by \clasp\
are shown in the last three columns,
here averaging over the instances finished within the time limit
in both preprocessing modes.

%
\begin{table}
\centering
	\newcommand{\HD}[1]{\multicolumn{1}{c}{#1}}
\newcommand{\HDD}[1]{\multicolumn{4}{c}{#1}}
\newcommand{\HDI}[1]{\multicolumn{2}{c}{#1}}
\renewcommand{\tabcolsep}{4pt}
\begin{tabular}{ 
	l
	r
	r@{/}r
	r@{/}r
	r@{(}r@{)/}r@{(}r
	r@{/}r
	r@{/}r
	r@{/}l
}
\toprule
        \textit{paranoid}     & \HD{$n$} & \HDI{variables}  & \HDI{constraints} & \HDD{time (t/o)}       & \HDI{choices}  & \HDI{conflicts} & \multicolumn{2}{@{}c}{answer sets}     \\
	\cmidrule{1-16}                                                                                                                             
	\cmidrule{1-1}                                                                                                                              
	\textit{easy}         &     20 &  6K &  69K       &   6K &  91K       &   1 &  0 &   9 &    0) &  35K &  1,932K &  22 &        27 &  66 &        192 \\
	\textit{difficult}    &     22 & 11K & 158K       &  10K & 180K       &   2 &  0 &  25 &    0) &  42K &    717K &  5K &        4K &  67 &         87 \\
	\textit{impossible}   &     14 & 36K & 404K       &  64K & 654K       &   6 &  0 &  98 &    0) &  90K &    992K &  7K &        5K &  58 &         81 \\
	\textit{debian-dudf}  &     18 & 40K & 189K       &  82K & 359K       &   6 &  0 &  40 &    0) & 232K &    953K &  2K &        1K & 220 &        116 \\
	\midrule                                                                                                                                    
	\textit{trendy}       & \HD{$n$} & \HDI{variables}  & \HDI{constraints} & \HDD{time (t/o)}       & \HDI{choices}  & \HDI{conflicts} & \multicolumn{2}{@{}c}{answer sets} \\                                                                                                                             
	\cmidrule{1-16}                                                                                                                              
	\textit{easy}         &     20 &  9K &  80K       &  11K & 121K       &   1 &  0 &  14 &    0) & 117K &  3,690K &  1K &        2K & 203 &        341 \\
	\textit{difficult}    &     22 & 21K & 175K       &  26K & 232K       & 155 & 11 & 196 &   12) & 279K &  3,057K & 26K &       28K & 270 &        400 \\
	\textit{impossible}   &     14 & 70K & 438K       & 136K & 782K       & 163 &  6 & 259 &   12) & 462K &  2,949K & 12K &       12K & 289 &        253 \\
	\textit{debian-dudf}  &     18 & 51K & 207K       & 111K & 432K       &  20 &  0 & 106 &    1) & 946K & 10,910K & 35K &       51K & 678 &        874 \\
\bottomrule
\end{tabular}

	\caption{Experiments assessing the impact
                 of preprocessing via Algorithm~\ref{algo:cudf2lp}
                 on  \aspcud's performance.\label{tab:experiments}}
\end{table}

With transitive closure computation enabled,
we observe a reduction of both variables and constraints by about one order of magnitude
(a bit less on the \textit{debian-dudf} class).
This can be explained by the fact that typical installations include
only a fraction of the available packages.
Furthermore, the reductions in size are greater
wrt.\ \textit{paranoid} objectives because they disregard recommendations,
which are considered in \textit{trendy} mode.
%
The solving times 
also reduce by one order of magnitude for \textit{paranoid},
yet less for the more difficult problems solved in \textit{trendy} mode;
however, eight more instances are solved in time with transitive closure computation enabled.
%
Interestingly, the numbers of conflicts and answer sets 
(taken only over instances that did not time out) are comparable.
This indicates that \clasp's optimization approach is able to focus on
relevant problem parts, even without a priori limitation to the transitive closure.
%
Nonetheless, the numbers of choices are much greater 
(again an order of magnitude) for whole package universes,
providing a clear indication of the benefits of limiting the scope of follow-up installations.
In fact, even when unnecessary variables and constraints do not render
a problem more difficult,
the solving time suffers from additional efforts spent on
assigning the variables and testing the constraints.

\section{Discussion}\label{sec:discussion}

We presented the workflow of the ASP-based Linux package configuration
tool \aspcud.
In particular, we detailed the preprocessing applied to
convert \cudf\ input to ASP facts suitable for the
ASP components of \aspcud.
Related approaches rely on conversions from \cudf\ to
Integer Linear Programming~\cite{micrue10a},
Maximum Satisfiability~\cite{jalymama11a}, or
Pseudo-Boolean Optimization~\cite{arbelymara10a}.
Although all conversions, including ours, closely follow the 
specification of \cudf~\cite{trezac09a} and differ primarily
in their target formats, there still are some differences
that deserve attention.
Unlike other package configuration tools,
\aspcud\ compiles \cudf\ input into ASP facts,
while constraints as well as optimization criteria on follow-up installations
are provided separately via general problem encodings.
In fact, \aspcud\ is equipped with several encoding variants
(selectable via switch `\code{-e}'),
although we here only detailed the most promising variants according to
our empirical investigations.
For another, the preprocessors of package configuration tools
trace indirections in view of arithmetic expressions (over versions),
package formulae, and virtual packages admitted in \cudf\
back to the (installable) packages underneath.
In our ASP fact format (cf.\ Figure~\ref{genfact}),
we however associate target sets~$P$ of package interdependencies
with identifiers $\mathit{id}_P$
in order to avoid unfolding steps upon fact generation.
To our knowledge,
the preprocessors of other package configuration tools perform such unfolding,
and it is an interesting (unresolved) question whether structural entities
of the form $\mathit{id}_P$ are rather beneficial or a handicap for search.
Regarding modeling in ASP
(cf.\ Figure~\ref{fig:configuration} and~\ref{fig:optimize}),
the consequent usage of identifiers $\mathit{id}_P$ helped to keep the
encodings concise and thus easy to maintain and modify.
Despite of the different input formats used in ASP and the solving components
of other package configuration tools, the principal approach of \aspcud's
preprocessor to limit the scope of follow-up installations is independent
of back-end solvers; however, an additional ``constraint formulator''
would be required for back-ends lacking general-purpose grounders.
Concerning subjects to future investigation, we speculate that further improvements of problem encodings or
the exploration of characteristic structures in Linux distributions (if any)
might boost the performance of package configuration tools, 
in addition to ongoing enhancements of their search engines.


\paragraph{Acknowledgments.}
This work was partly funded by DFG grant SCHA 550/8-2.
We are grateful to Daniel Le Berre
for useful discussions on the topic, 
to the {\small\textsf{mancoosi}} project team
for organizing MISC, 
and to the anonymous referees for helpful comments.





\begin{thebibliography}{10}
\providecommand{\bibitemdeclare}[2]{}
\providecommand{\urlprefix}{Available at }
\providecommand{\url}[1]{\texttt{#1}}
\providecommand{\href}[2]{\texttt{#2}}
\providecommand{\urlalt}[2]{\href{#1}{#2}}
\providecommand{\doi}[1]{doi:\urlalt{http://dx.doi.org/#1}{#1}}
\providecommand{\bibinfo}[2]{#2}

\bibitemdeclare{inproceedings}{arbelymara10a}
\bibitem{arbelymara10a}
\bibinfo{author}{J.~Argelich}, \bibinfo{author}{D.~{Le Berre}},
  \bibinfo{author}{I.~Lynce}, \bibinfo{author}{J.~Marques-Silva} \&
  \bibinfo{author}{P.~Rapicault} (\bibinfo{year}{2010}):
  \emph{\bibinfo{title}{Solving {L}inux Upgradeability Problems Using {B}oolean
  Optimization}}.
\newblock In \bibinfo{editor}{Lynce} \& \bibinfo{editor}{Treinen}
  \cite{lococo10}, pp. \bibinfo{pages}{11--22},
  \doi{10.4204/EPTCS.29.2}.

\bibitemdeclare{misc}{aspcud}
\bibitem{aspcud}
\emph{\bibinfo{title}{aspcud}}.
\newblock
  \bibinfo{howpublished}{\texttt{http://www.cs.uni-potsdam.de/wv/aspcud}}.

\bibitemdeclare{book}{baral02a}
\bibitem{baral02a}
\bibinfo{author}{C.~Baral} (\bibinfo{year}{2003}):
  \emph{\bibinfo{title}{Knowledge Representation, Reasoning and Declarative
  Problem Solving}}.
\newblock \bibinfo{publisher}{Cambridge University Press},
  \doi{10.1017/CBO9780511543357}.

\bibitemdeclare{book}{SATHandbook}
\bibitem{SATHandbook}
\bibinfo{editor}{A.~Biere}, \bibinfo{editor}{M.~Heule},
  \bibinfo{editor}{H.~{van Maaren}} \& \bibinfo{editor}{T.~Walsh}, editors
  (\bibinfo{year}{2009}): \emph{\bibinfo{title}{Handbook of Satisfiability}}.
\newblock
\bibinfo{publisher}{IOS Press}.

\bibitemdeclare{inproceedings}{gekakasc11b}
\bibitem{gekakasc11b}
\bibinfo{author}{M.~Gebser}, \bibinfo{author}{R.~Kaminski},
  \bibinfo{author}{B.~Kaufmann} \& \bibinfo{author}{T.~Schaub}
  (\bibinfo{year}{2011}): \emph{\bibinfo{title}{Multi-Criteria Optimization in
  Answer Set Programming}}.
\newblock In \bibinfo{editor}{J.~Gallagher} \& \bibinfo{editor}{M.~Gelfond},
  editors: {\sl \bibinfo{booktitle}{Technical Communications of the Twenty-seventh International
                  Conference on Logic Programming (ICLP'11)}},
  \bibinfo{publisher}{Leibniz International Proceedings in Informatics}, pp. 
  \bibinfo{pages}{1--10},
  \doi{10.4230/LIPIcs.ICLP.2011.1}.

\bibitemdeclare{inproceedings}{gekakasc11c}
\bibitem{gekakasc11c}
\bibinfo{author}{M.~Gebser}, \bibinfo{author}{R.~Kaminski},
  \bibinfo{author}{B.~Kaufmann} \& \bibinfo{author}{T.~Schaub}
  (\bibinfo{year}{2011}): \emph{\bibinfo{title}{Multi-Criteria Optimization in
  {ASP} and its Application to Linux Package Configuration}}.
\newblock In \bibinfo{editor}{{Le Berre}} \& \bibinfo{editor}{{Van Gelder}}
  \cite{pos11a}.
\newblock \bibinfo{note}{To appear}.

\bibitemdeclare{inproceedings}{gekakosc11a}
\bibitem{gekakosc11a}
\bibinfo{author}{M.~Gebser}, \bibinfo{author}{R.~Kaminski},
  \bibinfo{author}{A.~K{\"o}nig} \& \bibinfo{author}{T.~Schaub}
  (\bibinfo{year}{2011}): \emph{\bibinfo{title}{Advances in \textit{Gringo}
  Series 3}}.
\newblock In \bibinfo{editor}{J.~Delgrande} \& \bibinfo{editor}{W.~Faber},
  editors: {\sl \bibinfo{booktitle}{Proceedings of the Eleventh International
  Conference on Logic Programming and Nonmonotonic Reasoning (LPNMR'11)}}, 
  \bibinfo{publisher}{Springer}, pp.
  \bibinfo{pages}{345--351},
  \doi{10.1007/978-3-642-20895-9\_39}.

\bibitemdeclare{inproceedings}{gekanesc07a}
\bibitem{gekanesc07a}
\bibinfo{author}{M.~Gebser}, \bibinfo{author}{B.~Kaufmann},
  \bibinfo{author}{A.~Neumann} \& \bibinfo{author}{T.~Schaub}
  (\bibinfo{year}{2007}): \emph{\bibinfo{title}{Conflict-Driven Answer Set
  Solving}}.
\newblock In \bibinfo{editor}{M.~Veloso}, editor: {\sl
  \bibinfo{booktitle}{Proceedings of the Twentieth International Joint
  Conference on Artificial Intelligence (IJCAI'07)}}, \bibinfo{publisher}{AAAI
  Press/The MIT Press}, pp. \bibinfo{pages}{386--392}.

\bibitemdeclare{inproceedings}{jalymama11a}
\bibitem{jalymama11a}
\bibinfo{author}{M.~Janota}, \bibinfo{author}{I.~Lynce},
  \bibinfo{author}{J.~Marques-Silva} \& \bibinfo{author}{V.~Manquinho}
  (\bibinfo{year}{2011}): \emph{\bibinfo{title}{PackUp: Tools for Package
  Upgradability Solving}}.
\newblock In \bibinfo{editor}{{Le Berre}} \& \bibinfo{editor}{{Van Gelder}}
  \cite{pos11a}.
\newblock \bibinfo{note}{To appear}.

\bibitemdeclare{proceedings}{pos11a}
\bibitem{pos11a}
\bibinfo{editor}{D.~{Le Berre}} \& \bibinfo{editor}{A.~{Van Gelder}}, editors
  (\bibinfo{year}{2011}): \emph{\bibinfo{title}{Proceedings of the Second
  Workshop on Pragmatics of SAT (PoS'11)}}.
\newblock \bibinfo{note}{To appear}.

\bibitemdeclare{book}{KRHandbook}
\bibitem{KRHandbook}
\bibinfo{editor}{V.~Lifschitz}, \bibinfo{editor}{F.~{van Harmelen}} \&
  \bibinfo{editor}{B.~Porter}, editors (\bibinfo{year}{2008}):
  \emph{\bibinfo{title}{Handbook of Knowledge Representation}}.
\newblock \bibinfo{publisher}{Elsevier Science}.

\bibitemdeclare{proceedings}{lococo10}
\bibitem{lococo10}
\bibinfo{editor}{I.~Lynce} \& \bibinfo{editor}{R.~Treinen}, editors
  (\bibinfo{year}{2010}): \emph{\bibinfo{title}{Proceedings of the First
  International Workshop on Logics for Component Configuration (LoCoCo'10)}}.
  {\sl \bibinfo{series}{Electronic Proceedings in Theoretical Computer Science
  (EPTCS)}}~\bibinfo{volume}{29},
  \doi{10.4204/EPTCS.29}.

\bibitemdeclare{misc}{mancoosi}
\bibitem{mancoosi}
\emph{\bibinfo{title}{\textsf{\footnotesize\textup{mancoosi}} --- managing software complexity}}. 
\newblock \bibinfo{howpublished}{\texttt{http://www.mancoosi.org}}.

\bibitemdeclare{inproceedings}{micrue10a}
\bibitem{micrue10a}
\bibinfo{author}{C.~Michel} \& \bibinfo{author}{M.~Rueher}
  (\bibinfo{year}{2010}): \emph{\bibinfo{title}{Handling Software
  Upgradeability Problems with {MILP} Solvers}}.
\newblock In \bibinfo{editor}{Lynce} \& \bibinfo{editor}{Treinen}
  \cite{lococo10}, pp. \bibinfo{pages}{1--10},
  \doi{10.4204/EPTCS.29.1}.

\bibitemdeclare{misc}{potassco}
\bibitem{potassco}
\emph{\bibinfo{title}{potassco}}.
\newblock
\bibinfo{howpublished}{\texttt{http://potassco.sourceforge.net}}.

\bibitemdeclare{inproceedings}{syrjanen00a}
\bibitem{syrjanen00a}
\bibinfo{author}{T.~Syrj{\"a}nen} (\bibinfo{year}{2000}):
  \emph{\bibinfo{title}{Including Diagnostic Information in Configuration
  Models}}.
\newblock In \bibinfo{editor}{J.~Lloyd}, \bibinfo{editor}{V.~Dahl},
  \bibinfo{editor}{U.~Furbach}, \bibinfo{editor}{M.~Kerber},
  \bibinfo{editor}{K.~Lau}, \bibinfo{editor}{C.~Palamidessi},
  \bibinfo{editor}{L.~Pereira}, \bibinfo{editor}{Y.~Sagiv} \&
  \bibinfo{editor}{P.~Stuckey}, editors: {\sl \bibinfo{booktitle}{Proceedings
  of the First International Conference on Computational Logic (CL'00)}}, 
  \bibinfo{publisher}{Springer}, pp. \bibinfo{pages}{837--851},
  \doi{10.1007/3-540-44957-4\_56}.

\bibitemdeclare{techreport}{trezac09a}
\bibitem{trezac09a}
\bibinfo{author}{R.~Treinen} \& \bibinfo{author}{S.~Zacchiroli}
  (\bibinfo{year}{2009}): \emph{\bibinfo{title}{Common Upgradability
  Description Format ({CUDF}) 2.0}}.
\newblock \bibinfo{type}{Technical Report} \bibinfo{number}{003},
  \bibinfo{institution}{\cite{mancoosi}}. 

\end{thebibliography}
\end{document}